\documentclass[12pt]{article}

\usepackage{times}
\usepackage{amsmath}
\usepackage{amsfonts}
\usepackage{array}
\usepackage{microtype}
\usepackage{graphicx}
\usepackage{subfigure}
\usepackage{booktabs} 
\usepackage{hyperref}

\usepackage{algorithm}
\usepackage{algorithmic}

\usepackage[utf8]{inputenc} 
\usepackage[T1]{fontenc}    
\usepackage{hyperref}       
\usepackage{url}            
\usepackage{booktabs}       
\usepackage{amsfonts}       
\usepackage{nicefrac}       
\usepackage{microtype}      

\usepackage{array}  
\usepackage{booktabs}
\usepackage{natbib}

\title{Virtual-Taobao: Virtualizing Real-world Online Retail Environment for Reinforcement Learning} 

\author
{Jing-Cheng Shi$^{1,2}$, Yang Yu$^1$, Qing Da$^2$, Shi-Yong Chen$^1$, An-Xiang Zeng$^2$\\
\normalsize{$^1$National Key Laboratory for Novel Software Technology, Nanjing University, China}\\
\normalsize{$^2$Alibaba Group, China}
}

\date{}

\begin{document}
	\maketitle
	%
	%
	%
	%
	
	\begin{abstract}
		Applying reinforcement learning in physical-world tasks is extremely challenging. It is commonly infeasible to sample a large number of trials, as required by current reinforcement learning methods, in a physical environment. This paper reports our project on using reinforcement learning for better commodity search in Taobao, one of the largest online retail platforms and meanwhile a physical environment with a high sampling cost. Instead of training reinforcement learning in Taobao directly, we present our approach: 
		first we build \emph{Virtual Taobao}, a simulator learned from historical customer behavior data through the proposed GAN-SD (GAN for Simulating Distributions) and MAIL (multi-agent adversarial imitation learning), and then we train policies in Virtual Taobao with no physical costs in which ANC (Action Norm Constraint) strategy is proposed to reduce over-fitting. 
		In experiments, Virtual Taobao is trained from hundreds of millions of customers' records, and its properties are compared with the real environment. The results disclose that Virtual Taobao faithfully recovers important properties of the real environment. We also show that the policies trained in Virtual Taobao can have significantly superior online performance to the traditional supervised approaches. We hope our work could shed some light on reinforcement learning applications in complex physical environments.
	\end{abstract}
	
	\section{Introduction}
	
	With the incorporation of deep neural networks, Reinforcement Learning (RL) has achieved significant progress recently, yielding lots of successes  in games \citep{mnih2013playing,silver2016mastering,mnih2016asynchronous}, robotics \citep{kretzschmar2016socially}, natural language processing \citep{su2016line}, etc. However, there are few studies on the application of RL in physical-world tasks like large online systems interacting with customers, which may have great influence on the user experience as well as the social wealth. 
	
	Large online systems, though rarely incorporated with RL methods, indeed yearn for the embrace of RL. In fact, a variety of online systems involve the sequential decision making as well as the delayed feedbacks. For instance, an automated trading system needs to manage the portfolio according to the historical metrics and all the related information with a high frequency, and carefully adjusts its strategy through analyzing the long-term gains. Similarly, an E-commerce search engine observes a buyer's request, and displays a page of ranked commodities to the buyer, then updates its decision model after obtaining the user feedback to pursue revenue maximization. During a session, it keeps displaying new pages according to latest information of the buyer if he/she continues to browse. Previous solutions are mostly based on supervised learning. They are incapable of learning sequential decisions and maximizing long-term reward. Thus RL solutions are highly appealing, but was not well noticed only until recently \citep{kdd18:2,kdd18:1}.
	
	One major barrier to directly applying RL in these scenarios is that, current RL algorithms commonly require a large amount of interactions with the environment, which take high physical costs, such as real money, time from days to months,  bad user experience, and even lives in medical tasks. To avoid physical costs, simulators are often employed for RL training. Google's application of data center cooling~\citep{gao2014machine} demonstrates a good practice: the system dynamics are approximated by a neural network, and a policy is later trained in the simulated environment via some RL algorithms. In our project for commodity search in Taobao.com, we have the similar process: build a simulator, i.e, {Virtual Taobao}, then the policy can be trained offline in the simulator by any RL algorithm maximizing long-term reward. Ideally, the obtained policy would perform equally well in the real environment, or at least provides a good initialization for cheaper online tuning.
	
	However, different from approximating the dynamics of data centers, simulating the behavior of hundreds of millions of customers in a dynamic environment is much more challenging. We treat customers behavior data to be generated from customers' policies. Deriving a policy from data can be realized by existing \emph{imitation learning} approaches~\citep{schaal1999imitation,argall2009survey}. The behavior cloning (BC) methods~\citep{Pomerleau:Efficient} learn a policy mainly by supervised methods from the state-action data. 
	BC requires the i.i.d. assumption on the demonstration data that is unsatisfied in RL tasks. The inverse reinforcement learning (IRL) methods~\citep{ng2000algorithms} learn a reward function from the data, and a policy is then trained according to the reward function. IRL relaxes the i.i.d. assumption of the data, but still assumes that the environment is static. Note that the customer behavior data is collected under a fixed Taobao platform strategy. When we are training the platform strategy, the environment of the customers will change and thus the learned policy could fail. All the above issues make these method less  practical for building the Virtual Taobao.
	
	
	\begin{figure}[!t]
		\centering
		\includegraphics[width=1\linewidth]{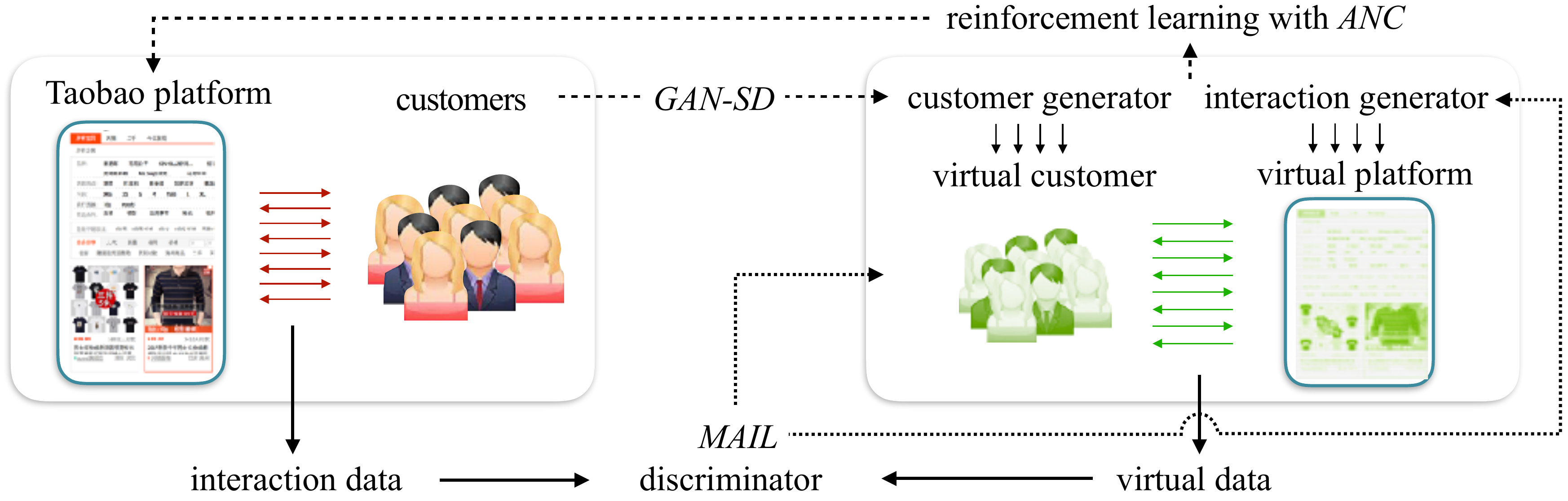}
		\vspace{-.5em}
		\caption{Virtual-Taobao architecture for reinforcement learning.}
		\label{vtaobao}
	\end{figure}

	In this work, we make Virtual Taobao through generating customers and generating interactions. Customers with search request, which has a very complex and widely spanned distribution, come into Taobao and trigger the platform search engine. We propose the GAN-for-Simulating-Distribution (GAN-SD) approach to simulate customers including their request.
	Since the original GAN methods often undesirably mismatch with the target distribution \citep{VEEGAN}, GAN-SD adopts an extra distribution constraint to generate diverse customers.
	To generate interactions, which is the key component of Virtual Taobao, we propose the Multi-agent Adversarial Imitation Learning ({MAIL}) approach.
	We could have directly invoked Taobao platform strategy in Virtual Taobao, which however makes a static environment that will not adapt to changes in the real environment. Therefore, MAIL learns the customers' policies and the platform policy simultaneously.
	In order to learn the two sides, MAIL follows the idea of {GAIL}~\citep{ho2016generative} using generative adversarial framework~\citep{goodfellow2014generative}. MAIL trains a discriminator to distinguish the simulated interactions from the real interactions; the discrimination signal feeds back as the reward to train the customer and platform policies for generating more real-alike  interactions.
	After generating customers and interactions, Virtual Taobao is built.
	As we find that a powerful algorithm may over fit to Virtual Taobao, which means it can do well in the virtual environment but poorly in the real, the proposed Action Norm Constraint (ANC) strategy can reduce such over-fitting.
	In experiments, we build Virtual Taobao from hundreds of millions of customers' records, and compare it with the real environment. Our results disclose that Virtual Taobao successfully reconstructs properties very close to the real environment. We then employ Virtual Taobao to train  platform policy for maximizing the revenue. Comparing with the traditional supervised learning approach, the strategy trained in Virtual Taobao achieves more than 2\% improvement of revenue in the real environment. 
	
	The rest of the sections present the background, the Virtual Taobao approach, the offline and online experiments, and the conclusion, in order.

	\section{Background}

	\subsection{Reinforcement Learning}
	
	Reinforcement learning (RL) solves sequential decision making problems through trial-and-error. We consider a standard RL formulation based on a Markov Decision Process (MDP).
	An MDP is described by a tuple $\left< \mathcal{S}, \mathcal{A}, \mathcal{T}, r, \gamma\right>$ where $\mathcal{S}$ is the observation space, $\mathcal{A}$ is the action space and $\gamma$ is the discount factor. 
	$\mathcal{T}:\mathcal{S}\times\mathcal{A}\rightarrow\mathcal{S}$ is the transition function to generate next states from the current state-action pair. $r:\mathcal{S}\times\mathcal{A}\rightarrow \mathbb{R}$ denotes the reward function. 
	At each timestep $t$, the RL agent observes a state $s_t \in \mathcal{S}$ and chooses an action $a_t \in \mathcal{A}$ following a policy $\pi$. 
	Then it will be transfered into a new state $s_{t+1}$ according to $\mathcal{T}(s_t,a_t)$ and receives an immediate reward $r_t = \mathcal{R}(s_t,a_t)$. 
	$R_t = \sum_{k=0}^{\infty} \gamma^t r_{t+k}$ is the discounted, accumulated reward with the discount factor $\gamma$. 
	The goal of RL is to learn a policy $\pi : \mathcal{S}\times\mathcal{A} \rightarrow [0,1]$ that solves the MDP by maximizing the expected discounted return, i.e., 
	$\pi^* = \arg\max E[\,R_t\,]$.

	\subsection{Imitation Learning}
	
	It is apparently a more effective task to act following by a teacher rather than learning a policy from scratch.
	In addition, the manual-designed reward function in RL may not be the real one \citep{Farley:mechanical,Hoyt:Gait} and a tiny change of reward function in RL may result in a totally different policy. In imitation learning, the agent is given trajectory samples from some expert policy, and infers the expert policy.
	
	\textbf{Behavior Cloning} \& \textbf{Inverse Reinforcement Learning} are two traditional approaches for imitation learning: behavior cloning learns a policy as a supervised learning problem over state-action pairs from expert trajectories \citep{Pomerleau:Efficient}; and inverse reinforcement learning finds a reward function under which the expert is uniquely optimal \citep{Stuart:Learning}, and then train a policy according to the reward.
	
	While behavior cloning methods are powerful for one-shot imitation \citep{duan2017one}, it needs large training data to work even on non-trivial tasks, due to compounding error caused by covariate shift \citep{ross2010efficient}. It also tends to be brittle and fails when the agent diverges too much from the demonstration trajectories \citep{ross2011reduction}. On the other hand, inverse reinforcement learning finds the reward function being optimized by the expert, so compounding error, a problem for methods that fit single-step decisions, is not an issue. 
	IRL algorithms are often expensive to run because they need reinforcement learning in a inner loop. Some works have focused on scaling IRL to large environments \citep{finn2016guided}.
	
	\subsection{Generative Adversarial Networks}
	Generative adversarial networks ({GAN}s) \citep{goodfellow2014generative} and its variants are rapidly emerging unsupervised machine learning techniques. They are implemented by a system of two neural networks contesting with each other in a zero-sum game framework. They train a discriminator $D$ to maximize the probability of assigning the correct labels to both training examples and generated samples, and a generator $G$ to minimize the classification accuracy according to $D$. The discriminator and the generator are implemented by neural networks, and are updated alternately in a competitive way.
	Recent studies have shown that GANs are capable of generating faithful real-world images \citep{CycleGAN}, implying their applicability in modeling complex distributions.
	
	\textbf{Generative Adversarial Imitation Learning} 
	\citep{ho2016generative} was recently proposed to overcome the brittleness of behavior cloning as well as the expensiveness of inverse reinforcement learning using GAN framework.
	GAIL allows the agent to interact with the environment and learns the policy by RL methods while the reward function is improved during training. Thus the RL method is the generator in the GAN framework.
	GAIL employs a discriminator $D$ to measure the similarity between the policy-generated trajectories and the expert trajectories. 
	
	GAIL is proven to achieve the similar theoretical and empirical results as IRL and GAIL is more efficient.
	GAIL has become a popular choice for imitation learning \citep{kuefler2017imitating} and there already exist model-based \citep{baram2016model} and third-person \citep{stadie2017third} extensions. 
	
	
	\section{Virtual Taobao}
	
	\subsection{Problem Description}
	

	Commodity search is the core business in Taobao, one of the largest retail platforms. 
	Taobao can be considered as a system where the search engine interacts with customers. 
	The search engine in Taobao deals with millisecond-level responses to billions of commodities, while the customers' preference of commodities are also rich and diverse.
	From the engine's point of view, Taobao platform works as the following.
	A customer comes and sends a search request to the search engine. Then the engine makes an appropriate response to the request by sorting the related commodities and displaying the page view (PV) to the customer. The customer gives the feedback signal, e.g. buying, turning to the next page, and leaving, according to the PVs as well as the buyer's own intention. The search engine receives the signal and makes a new decision for the next PV request. The business goal of Taobao is to increase sales by optimizing the strategy of displaying PVs. As the feedback signal from a customer depends on a sequence of PVs, it's reasonable to consider it as a multi-step decision problem rather than a one-step supervised learning problem.
	
	
	Considering the search engine as an agent, and the customers as an environment, commodity search is a sequential decision making problem. It is reasonable to assume customers can only remember a limited number, $m$, of the latest PVs, which means the feedback signals are only influenced by $m$ historical actions of the search agent. Let $a\in \mathcal{A}$ denote the action of the search engine agent and $F_a$ denote the feedback distribution from customers at $a$ , given the historical actions from the engine $a_0,a_1,...,a_{n-1}$, we have
	$
	F_{a_n|a_{n-1}, a_{n-2}, ..., a_{0}} = F_{a_n|a_{n-1}, ..., a_{n-m}}
	$.
	
	On the other hand, the shopping process for a customer can also be viewed as a sequential decision process. As a customer's feedback is influenced by the last $m$ PVs, which are generated by the search engine and are affected by the last feedback from the customer. The customers' behaviors also have the Markov property. The process of developing the shopping preference for a customer can be regarded as a process optimizing his shopping policy in Taobao.
	
	\begin{figure}[!h]
		\centering
		\includegraphics[width=0.9\linewidth]{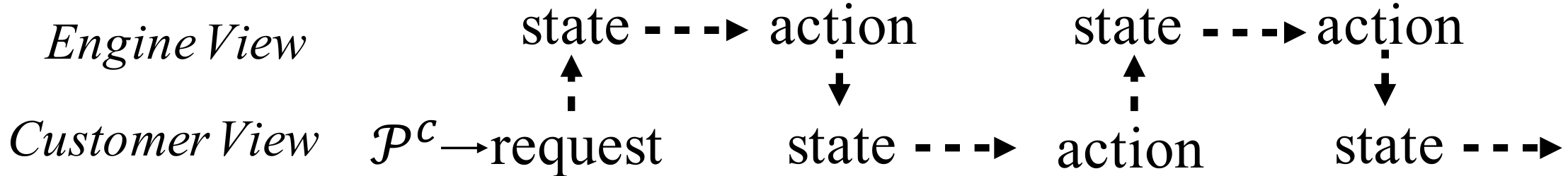}\\[2em]
		\begin{tabular}{c|c|c}	
			\hline
			~&Engine view&Customer view\\
			\hline
			State & customer feature and request& customer feature, engine action and PV info\\ 
			\hline
			Action & parametermized as a vector in $\mathbb{R}^d$ &buy, turn page or leave\\
			\hline
		\end{tabular}
		\caption{Taobao search in engine view and in customer view}
		\label{fig_customer_engine}
	\end{figure}
	
	The engine and customers are the environments of each other. Figure \ref{fig_customer_engine} shows the details. In this work, PV info only contains the page index.
	
	We use $\mathcal{M}\!=\!\left<\,\mathcal{S},\mathcal{A},\mathcal{T},\mathcal{R},\pi\,\right>$ and $\mathcal{M}^c\!=\!\left<\,\mathcal{S}^c,\mathcal{A}^c,\mathcal{T}^c, \mathcal{R}^c, \pi^c\,\right>$ to represent the decision process for engine and customers. Note that the customers with requests come from the real environment $\mathcal{P}^c$.
	
	For the engine, the state remains the same if the customer turn to the next page. The state changes if the customer sends another request or leaves Taobao. 
	For the customer, As $\mathcal{S}^c=\mathcal{S}\times\mathcal{A}\times{\mathbb{N}}$, where $\mathbb{N}$ denotes the page index space, $s^c \in \mathcal{S}^c$ can be written as $\left<s,a,n\right>\in\mathcal{S}\times\mathcal{A}\times{\mathbb{N}}$. Formally:
	\begin{equation*}\label{eq_transition}
	\mathcal{T}(s,a)\!=\!
	\begin{cases}
	s, \text{  if } a^c\!=\!\text{turn page}\\
	s' \sim \mathcal{P}^c,\text{  otherwise} 
	\end{cases} ~~~
	\mathcal{T}^c(s^c,a^c)\!=\!
	\begin{cases}
	\left<s',\pi(s'),0\right>, s'\sim \mathcal{P}^c \text{ if } a^c\!=\!\text{leave}\\
	\left<s, a, n\!+\!1\right>,\text{if } a^c\!=\!\text{turn page}\\
	\mbox{terminates},\text{if } a^c\!=\!\text{buy, or } n\!>\!\text{MaxIndex}
	\end{cases}
	\end{equation*}
	For the engine, if the customer buy something, we give the engine a reward of 1 otherwise 0. For customers, the reward function is currently unknown to us. 
	\subsection{GAN-SD: Generating Customers}
	To build the Virtual Taobao, we need to firstly generate the customer, i.e. sample a user $U^c$ that includes its request from $\mathcal{P}^c$ to trigger the interaction process. We aims at constructing a sample generator to produce similar customers with that of the real Taobao. It's known that GANs are designed to generate samples close to the original data and achieves great success for generating images. We thus employ GANs to tightly simulate the real request generator. 
	
	However, we find that GANs empirically tends to generate the most frequent occurring customers. To generate a distribution rather than a single instance, we propose Generative Adversarial Networks for Simulating Distribution (GAN-SD), as in Algorithm \ref{alg_gan}.  
	Similar to GANs, GAN-SD maintains a generator $G$ and a discriminator $D$. 
	The discriminator tries to correctly distinguish the generated data from the training data by maximizing the following objective function 
	$$ E_{p_{x\sim\mathcal{D}}}[\log D(x)] + E_{p_{z\sim\mathcal{G}}}[\log (1-D(G(z)))],$$ 
	while the generator is updated to maximize the following objective function
	$$E_{p_{x\sim\mathcal{D}}, p_{z\sim\mathcal{G}}} [  D(G(z)) + \alpha \mathcal{H}(V(G(z))) - \beta KL (V(G(z))||V(x))]. $$
	$G(z)$ is the generated instance from the noise sample $z$. $V(\cdot)$ denotes some variable associated with the inner value. In our implementation, $V(\cdot)$ is the customer type of the instance. $\mathcal{H}(V(G(z)))$ denotes the entropy of the variable from the generated data, which is used to make a wider distribution. $KL (V(G(z))\|V(x))$ is the $KL$-divergence between the variables from the generated and training data, which is used to guide the generated distribution by the distribution in the training data. With the entropy and $KL$ divergence constraints, GAN-SD learns a generator with more guided information form the real data, and can generate much better distribution than the original GAN.
	
	\begin{algorithm}[!t]
		\caption{GAN-SD} 
		\label{alg_gan}
		\begin{algorithmic}[1] 
			\STATE \textbf{Input:} Real data distribution $\mathcal{P}_\mathcal{D}$
			\STATE Initialize training variables $\theta_D, \theta_G$
			\FOR{$i = 0,1,2...$} 
			\FOR {$k$ steps}
			\STATE Sample minibatch from noise prior $p_{\mathcal{G}}$
			and real dataset $p_{\mathcal{D}}$
			\STATE Update the generator by gradient:
			\vspace{-0.4em}$$ {\nabla}_{\theta_G} {E}_{p_{x\sim\mathcal{G}}, p_{x\sim\mathcal{D}}} [  D(G(z)) + \alpha \mathcal{H}(G(z)) - \beta KL (G(z)||x)] $$
			\vspace{-1.5em}\ENDFOR
			\STATE Sample minibatch from noise prior $p_{\mathcal{G}}$ and real dataset $p_{\mathcal{D}}$
			\STATE Update the discriminator by gradient:
			\vspace{-.4em}$$ \nabla_{\theta_D} E_{p_{x\sim\mathcal{D}}}[\log D(x)] + E_{p_{x\sim\mathcal{G}}}[\log (1-D(G(z)))]$$                    
			\vspace{-1.5em}\ENDFOR 
			\STATE \textbf{Output:} Customer generator $G$
		\end{algorithmic} 
	\end{algorithm}
	
	\subsection{MAIL: Generating Interactions}
	We generate interactions between the customers and the platform, which is the key of Virtual Taobao, through simulating customer policy.
	We accomplish this by proposing the Multi-agent Adversarial Imitation Learning (MAIL) approach following the idea of GAIL.
	
	Different from GAIL that trains one agent policy in a static environment, MAIL is a multi-agent approach that trains the customer policy as well as the engine policy. In such way, the learned customer policy is able to generalize with different engine policies.
	By MAIL, to imitate the customer agent policy $\pi^c$, we should know the environment of the agent, which means we also need to imitate $\mathcal{P}^c$ and $\mathcal{T}^c$, together with reward function $\mathcal{R}^c$. 
	The reward function is designed as the non-discriminativeness of the generated data and the historical state-action pairs. 
	The employed RL algorithm will maximize the reward, implying generating indistinguishable data.
	
	\begin{algorithm} 
		\caption{MAIL} 
		\label{alg_mail}
		\begin{algorithmic}[1] 
			\STATE \textbf{Input:} Expert trajectories $\tau_e$, customer distribution $\mathcal P^c$
			\STATE Initialize variables $\kappa, \sigma, \theta$
			\FOR{$i = 0,1,2...,I$} 
			\FOR{$j = 0,1,2...,J$} 
			\STATE $\tau_j = \emptyset, \, s \sim   \mathcal{P}^c, a\sim\pi_\sigma(s,\cdot), s^c=\left<s, a\right>$
			\WHILE{NOT TERMINATED}
			\STATE sample $a^c \sim \pi(s^c, \cdot)$, 
			add $(s^c, a^c)$ to $\tau_j$, 
			generate $s^c \sim \mathcal{T}^c_\sigma(s^c, a^c|\mathcal P^c)$
			\ENDWHILE
			\ENDFOR
			\STATE Sample trajectories $\tau_g$ from ${\tau}_{\,\mbox{\tiny{0}} \sim {\tiny{J}}}  $
			\STATE Update $\theta$ to in the direction to maximize
			\vspace{-.4em}
			\[
			E_{\tau_g}[\log(\mathcal{R}^c_\theta(s,a))]+E_{\tau_e}[\log(1 - \mathcal{R}^c_\theta(s,a))]
			\]
			\vspace{-1.5em}\STATE Update ${\kappa, \sigma}$ by optimizing $\pi^c_{\kappa, \sigma}$ with RL in $\mathcal{M}^c$
			\ENDFOR 
			\STATE \textbf{Output:} The customer agent policy $\pi^c$
		\end{algorithmic} 
	\end{algorithm}
	
	However, training the two policies, e.g., iteratively, could results in a very large search space, and thus poor performance. Fortunately, we can optimize them jointly. 
	We parameterize the customer policy  $\pi^c_\kappa$ by $\kappa$, the search engine policy $\pi_\sigma$ by $\sigma$, and the reward function $\mathcal{R}_\theta$ by $\theta$. We also denote $\mathcal{T}^c$ as $\mathcal{T}^c_\sigma$.
	Due to the relationship of the two policies:
	$$
	\pi^c(s^c,a^c) = \pi^c(\left<s,a,n\right>, a^c)=\pi^c(\left<s,\pi(s,\cdot),n\right>, a^c)
	$$
	which shows that, 
	given the engine policy, the joint policy $\pi^c_{\kappa, \sigma}$ can be viewed as  $S\times\mathbb{N}$ to $A^c$. 
	As $\mathcal{S}^c\!=\!\mathcal{S}\times\mathcal{A}\times\mathbb{N}$, we still consider $\pi^c_{\kappa, \sigma}$ as a mapping from $\mathcal{S}^c$ to $\mathcal{A}^c $ for convenience.
	The formalization of joint policy brings the chance to simulate $\pi$ and $\pi^c$ together. 

	We display the procedure of MAIL in Algorithm \ref{alg_mail}. 
	To run MAIL, we need the historical trajectories $\tau_e$ and a customer distribution $\mathcal P^c$, which is required by $\mathcal{T}^c_\sigma$.
	In this paper,  $\mathcal P^c$ is learned by GAN-SD in advance. 
	After initializing the variables, we start the main procedure of MAIL:
	In each iteration, 
	we collect trajectories during the interaction between the customer agent and the environment (line 4-9).
	Then we sample from the generated trajectories and optimize reward function by a gradient method (line 10-11).
	Then, $\kappa, \tau$ will be updated by optimizing the joint policy $\pi^c_{\kappa, \sigma}$ in $\mathcal{M}^c$ with a RL method (line 12).
	When the iteration terminates, MAIL returns the customer agent policy $\pi^c$.

	After simulating the distribution $\mathcal P^c$ and policy $\pi^c$, which means we know how customers are generated and how they response to PVs, Virtual Taobao is build. We can generate the interactions by deploy the engine policy to the virtual environment.
	
	\subsection{ANC: Reduce Over-fit to Virtual Taobao}
	We observe that if the deployed policy is similar to the historical policy, Virtual Taobao tends to behave more similarly than if the two policies are completely different. However, similar policy means little improvement.
	A powerful RL algorithm, such as TRPO, can easily train an agent to over-fit Virtual Taobao which means it can performs well in the virtual environment but poorly in the real environment. 
	Actually, we need to trade off between the accuracy and the improvement. 
	However, as the historical engine policy is unavailable in practice, we can not measure the distance between a current policy and the historical one. Fortunately, we note that the norms of most historical actions are relatively small, we propose the Action Norm Constraint (ANC) strategy to control the norm of taken actions. That is, when the norm of the taken action is larger than the norms of most historical actions, we give the agent a penalty. Specifically, we modified the original reward function as follow:
	\vspace{-.4em}\begin{equation}\label{eq_modified_reward}
	r'(s,a)=\frac{r(s,a)}{1 + \rho\,\max\{|\!|a |\!|-\mu,\,0\} }
	\end{equation}
	
	\section{Empirical Study}
	
	\subsection{Experiment Setting}
	To verify the effect of Virtual Taobao, we use following measurements as indicators.
	
	\begin{itemize}
	\item \textbf{Total Turnover (TT)} The value of the commodities sold.
	\item \textbf{Total Volume (TV)} The amount of the commodities sold.
	\item \textbf{Rate of Purchase Page (R2P)} The ratio of the number of PVs where purchase takes place.
    \end{itemize}

	All the measurements are adopted in online experiments. In offline experiments we only use R2P as we didn't predict the customer number and commodity price in this work.
	For the convenience of comparing these indicators between the real and virtual environment, we deployed a random engine policy in the real environments (specifically, an online A/B test bucket in Taobao) in advance and collect the corresponding trajectories as the historical data (about 400 million records). 
	Note that, we have no assumption of the engine policy which generates the data, i.e., it could be any unknown complex model. 
	
	
	We simulate the customer distribution $\mathcal P^c$ by GAN-SD with $\alpha=\beta=1$. 
	Then we build Virtual Taobao by MAIL. The RL method in MAIL we used is TRPO.
	All of function approximators in this work are implemented by multi-layer perceptions. 
	Due to the resource limitation, we can only compare two policies at the same time in online experiments. 
	%
	\subsection{On Virtual Taobao Properties}
	

	\subsubsection{Proportion \& R2P over Features}
	
\begin{figure*}[t]
		\centering
		\begin{minipage}{0.95\linewidth}
			\centering
			\includegraphics[width=\linewidth]{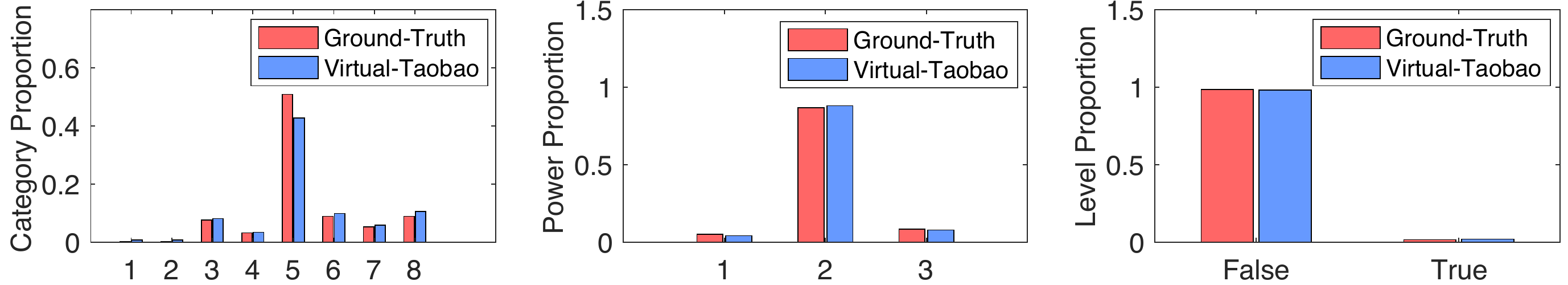}\vspace{-1em}
			\caption{The customer distributions between Taobao and the Virtual Taobao.}
			\label{num_vs_feature}
		\end{minipage}\\[2em]
		
		\begin{minipage}{0.95\linewidth}
			\centering
			\vspace{.5em}
			\includegraphics[width=\linewidth]{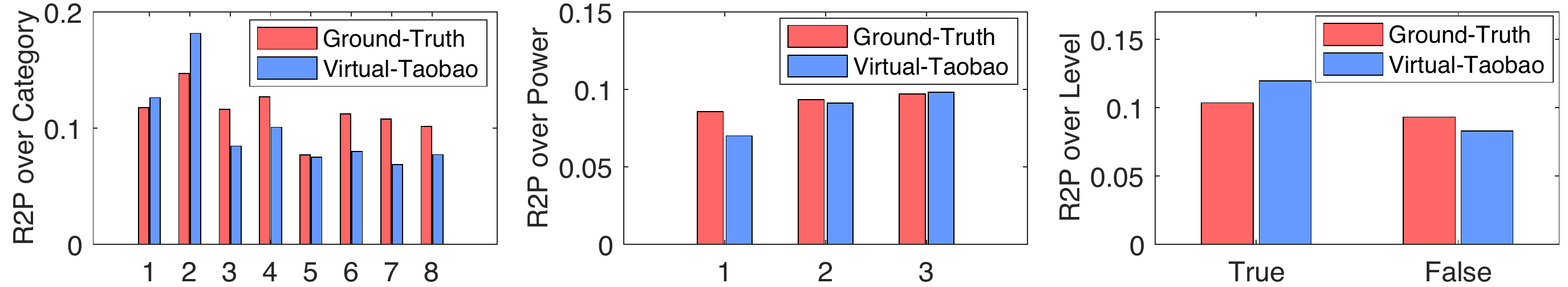}\vspace{-1em}
			\caption{The R2P distributions between Taobao and the Virtual Taobao.}
			\label{rate_vs_feature}
		\end{minipage}
	\end{figure*}

	The proportion of different customers is a basic criterion for testing Virtual Taobao.
	We generate 1,000,000 customers by $\mathcal P^c$ and calculate the proportions over three features respectively, i.e. query category (from one to eight), purchase power (from one to three) and high level indicator (True or False). 
	We compare the result with the ground truth, i.e., proportions in Taobao.
	Figure \ref{num_vs_feature} indicates that distribution is similar in the virtual and real environments.
	
	People have different preference of consumption which reflects in R2P.
	We report the influence of customer features on the R2P in Virtual Taobao and compare it with the results in the real. 
	As shown in Figure \ref{rate_vs_feature}, the results in Virtual Taobao are quite similar with the ground truth.
	
	\subsubsection{R2P over Time}
	R2P of the customers varies with time in Taobao, thus Virtual Taobao should have the similar property. 
	Since our customer model is independent of time, we divide the one-day historical data into 12 parts in order of time to simulate process of R2P changing over time. We train a virtual environment on every divided dataset independently. Then we deploy the same historical engine policy, i.e. the random policy, in the virtual environments respectively. We report the R2Ps in the virtual and the real environment. Figure \ref{rate_vs_time} indicates  Virtual Taobao can reflect the trend of R2P over time.
	
	\subsection{Reinforcement Learning in Virtual Taobao}

	\subsubsection{Generalization Capability of ANC}
	Virtual Taobao should have generalization ability as it is built from the past but serves for the future. 
	
\begin{figure}[t]
		\centering
		\begin{minipage}{0.7\linewidth}
			\centering
			\includegraphics[width=\linewidth]{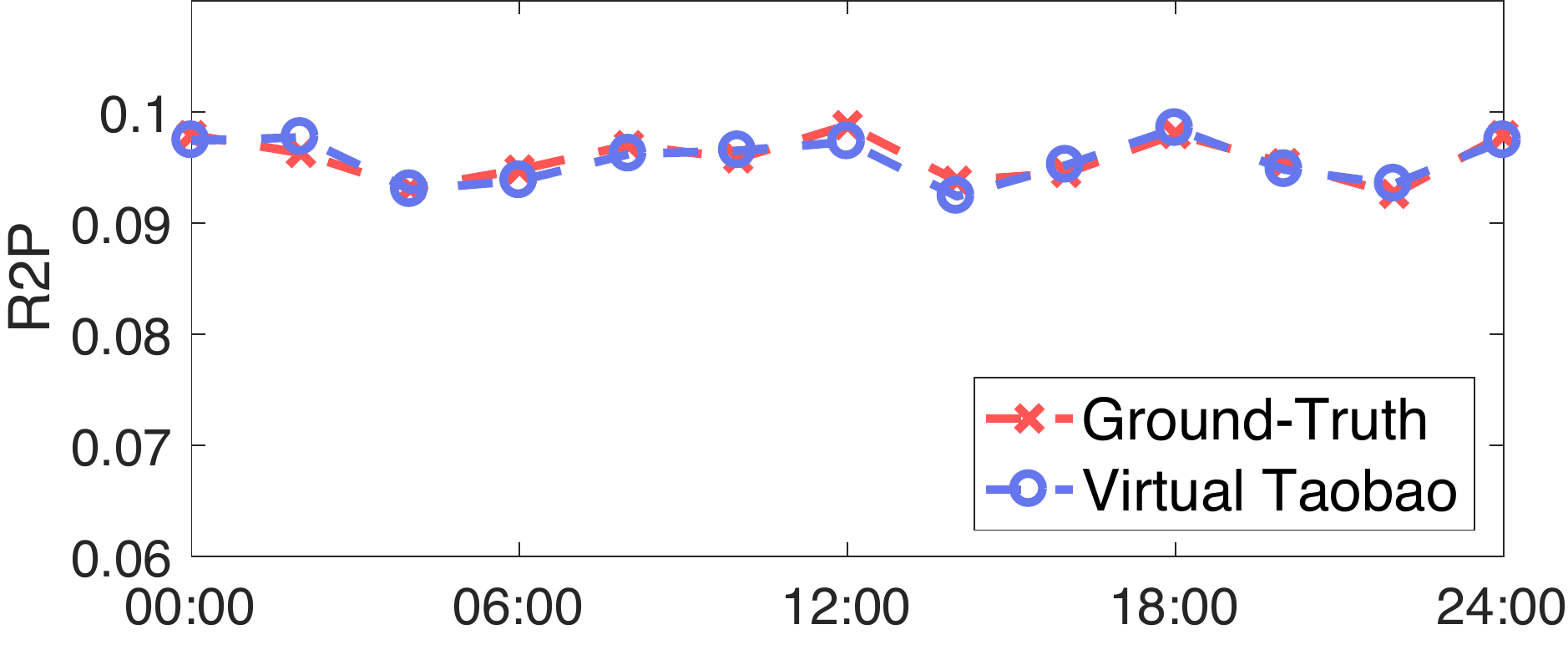}\vspace{-1.5em}
			\caption{R2P over time}
			\label{rate_vs_time}
		\end{minipage}\\[1em]
		\begin{minipage}{0.7\linewidth}
			\centering
			\includegraphics[width=\linewidth]{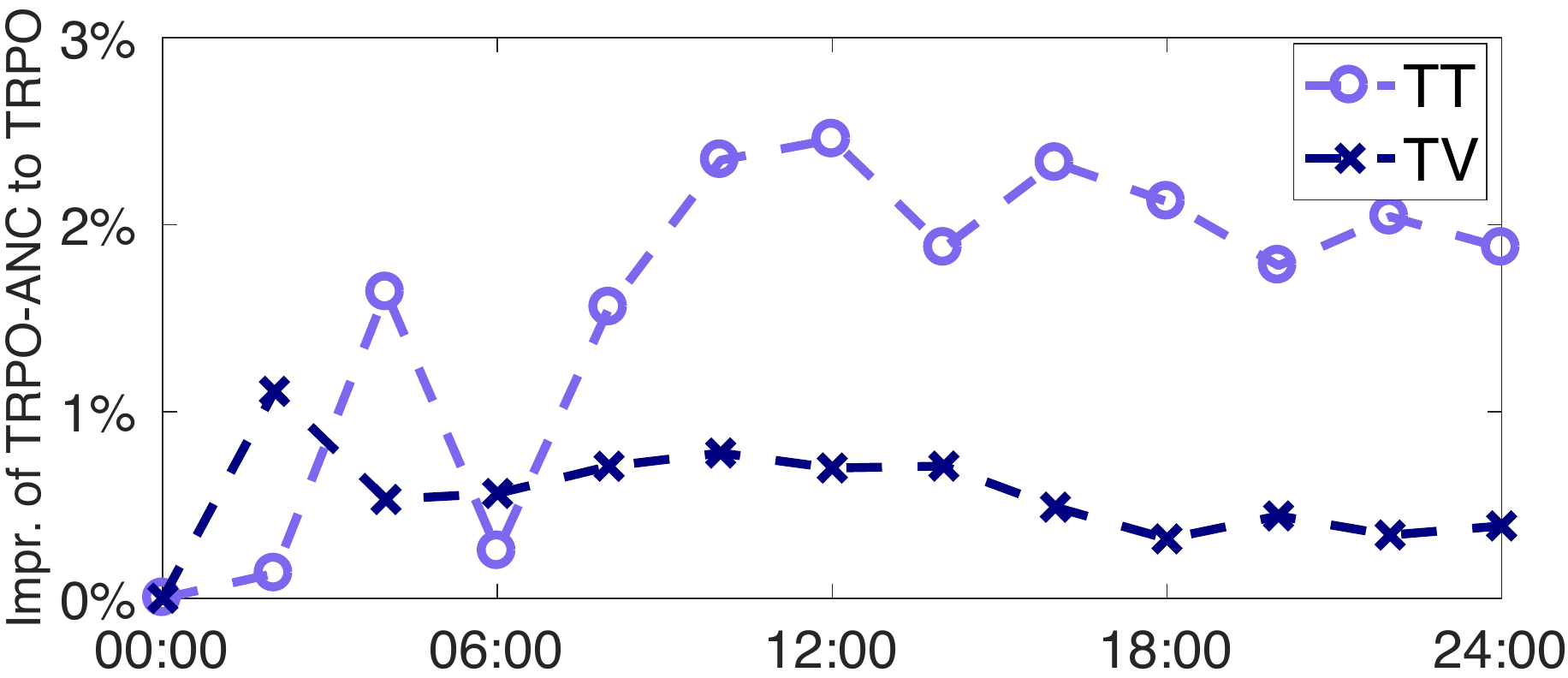}
			\vspace{-1.3em}\caption{Impr. of TRPO-ANC to TRPO\vspace{-1.3em}}
			\label{fig_pi^cons}
		\end{minipage}
	\end{figure}

	Firstly, we will show that the ANC strategy's ability on generalization.
	We train TRPO and TRPO-ANC, in which $\rho=1$ and $\mu=0.01$, in Vitual Taobao, and compare the results in real Taobao. Figure \ref{fig_pi^cons} shows the TT and TV increase of TRPO-ANC to TRPO. TRPO-ANC policy is always better than TRPO policy which indicates that the ANC strategy can reduce over-fitting to Virtual Taobao.
	
	\subsubsection{Generalization Capability of MAIL}
	
	Then, we'll verify the generalization ability of MAIL. We build a virtual environment by MAIL from one-day's data and build another 3 environments by MAIL whose data is from one day, a week and a month later. 
	We run TRPO-ANC in the first environment and deploy the result policy in other environments and see the decrease of R2P.
	We repeat the same process except that we replace MAIL with a behavior cloning method (BC). Figure \ref{fig_generalization} shows the R2P improvement of the policy trained in the two Virtual Taobao to a random policy. The R2P drops faster in the BC environment. The policy in BC environment even performs worse than then random policy after a month.

		\begin{table}[t]
			\centering
	 		\includegraphics[width=0.7\linewidth]{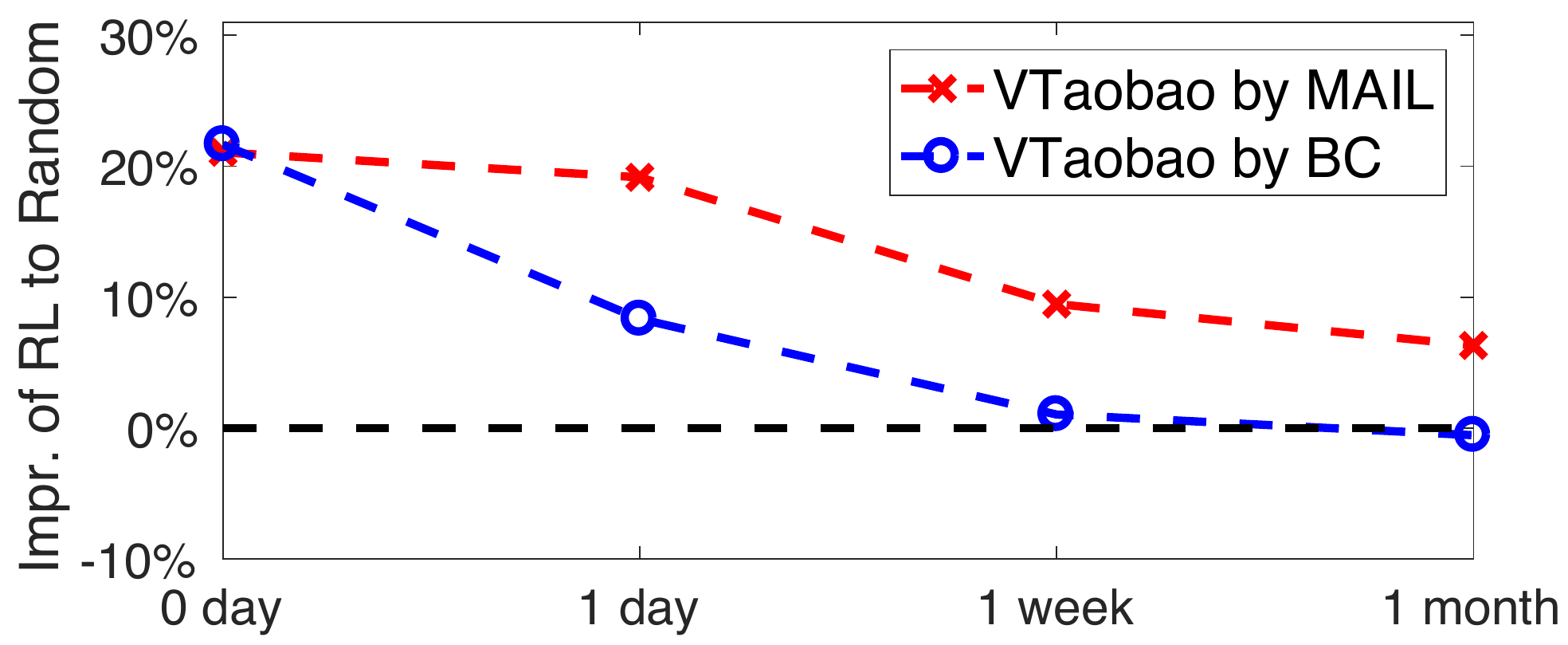}
	 		\caption{Generalization capability of Virtual Taobao}
	 		\label{fig_generalization}
	 	\end{table}
	
	\subsubsection{Online Experiments}
	We compare the policy generated by RL methods on Virtual Taobao (\emph{RL+VTaobao}) with supervised learning methods on the historical data (\emph{SL+Data}).
	Note that Virtual Taobao is constructed only from the historical data.
	To learn a engine policy using supervised learning method, we divide the historical data $S$ into two parts: $S_1$ contains the records with purchase and $S_0$ contains the other. 
	
	The first benchmark method, denoted by SL$_1$, is a classic regression model. 
	\begin{equation*}\label{eq_sl1}
	\pi_{SL_1}^{*} = \text{argmin}_{\pi} \frac{1}{|S_1|}\sum_{(s,a)\in S_1} \vert \pi(s)-a \vert ^ 2 
	\end{equation*}
	
	The second benchmark SL$_2$  modified the loss function and the optimal policy is defined as follow
	\begin{equation*}\label{eq_sl2}
	\pi_{SL_2}^{*} = \text{argmin}_{\pi}\frac{1}{|S_1|}\sum_{(s,a)\in S_1} \vert \pi(s)-a \vert ^ 2
	- \frac{\lambda_1}{|S_2|}\sum_{(s,a)\in S_2} \vert \pi(s) - a \vert ^ 2 + \frac{\lambda_2}{|S|}\sum_{(s,a)\in S} \vert\pi(s)\vert ^ 2
	\end{equation*}

	\begin{figure}[t]
		\centering
		\includegraphics[width=\linewidth]{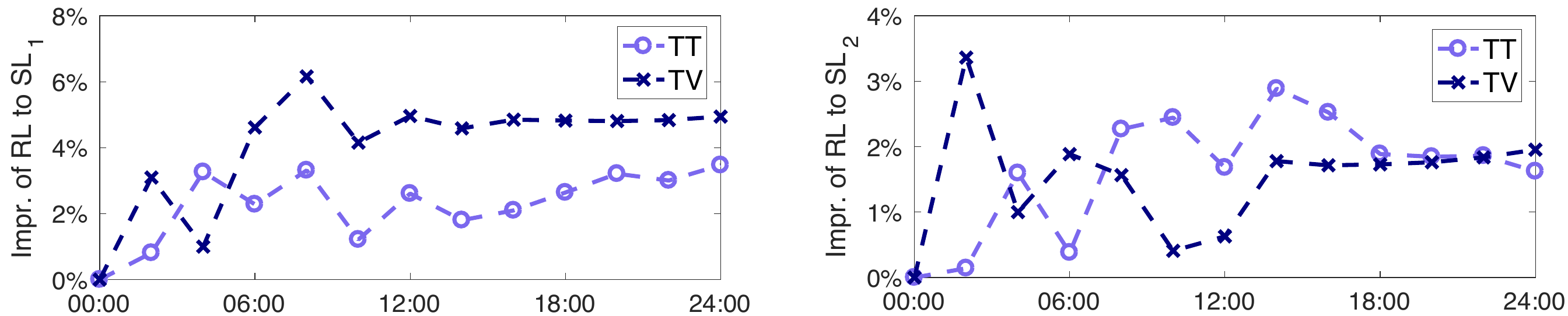}
		\caption{Improvement of {RL}+VTaobao to {SL}$_1$+data \&  {SL}$_2$+data in Taobao}
		\label{fig_rl_vs_sl}
	\end{figure}
	
	In our experiment, $\lambda_1 = 0.3$ and $\lambda_2 = 0.001$.
	As we can only compare two algorithms online in the same time due to the resource limitation, we report the results of RL v.s. SL$_1$ and RL v.s. SL$_2$ respectively. The results of TT and TV improvements to SL policies are shown in Figure \ref{fig_rl_vs_sl} . The R2P in the real environment of SL$_1$, SL$_2$ and RL are 0.096, 0.098 and 0.101 respectively. The \emph{RL+VTaobao} is always better than \emph{SL+Data}.

	\section{Conclusion}
	To overcome the high physical cost of training RL for commodities search in Taobao, we build the Virtual Taobao simulator which is trained from historical data by GAN-SD and MAIL. The empirical results have verified that it can reflect properties of the real environment faithfully. We then train better engine policies with proposed  ANC strategy in the Virtual Taobao, which has been shown to have better real environment performance than the traditional SL approaches. Virtualizing Taobao is very challenging. We hope this work can shed some light for applying RL to complex physical tasks.
	
	\small
	\bibliographystyle{icml2018}
	\bibliography{icml18}
\end{document}